%% file: 0_main.tex
\documentclass[sigconf]{acmart}

\AtBeginDocument{%
  \providecommand\BibTeX{{%
    \normalfont B\kern-0.5em{\scshape i\kern-0.25em b}\kern-0.8em\TeX}}}

\setcopyright{acmcopyright}
\copyrightyear{2024}
\acmYear{2024}
\usepackage{multirow}
\usepackage{multicol}
\usepackage{setspace}
\usepackage{appendix}
\usepackage{makecell}
\usepackage{rotfloat}
\usepackage{longtable}
\usepackage{pdflscape}
\usepackage{makecell}
\usepackage{color}
\usepackage{enumitem}
\usepackage{tikz}
\usepackage{tablefootnote}
\usetikzlibrary{mindmap}
\usepackage{balance}

\usetikzlibrary{shapes,positioning,shadows,trees}

\tikzset{
  basic/.style  = {draw, text width=5cm, drop shadow, font=\sffamily, rectangle},
  root/.style   = {basic, rounded corners=2pt, thin, align=center, fill=g},
  level 2/.style = {basic, rounded corners=2pt, thin,align=center, fill=p,
  text width=8em},
  level 3/.style = {basic, rounded corners=2pt, thin,align=center, fill=y,
  text width=8em},
  level 4/.style = {basic, thin, align=left, fill=b, text width=10 em}
}
\definecolor{g}{RGB}{217,244,212}
\definecolor{p}{RGB}{237,203,201}
\definecolor{b}{RGB}{237,244,252}
\definecolor{y}{RGB}{250,235,215}
\definecolor{lightgreen}{RGB}{44,160,44}
\setlist[itemize]{leftmargin=*}

\begin{document}

\title{A Survey on RAG Meeting LLMs: Towards Retrieval-Augmented Large Language Models}

\author{Wenqi Fan}
\email{wenqifan03@gmail.com}
\affiliation{
  \institution{The Hong Kong Polytechnic University, HK SAR}}
  
\author{Yujuan Ding}
\email{dingyujuan385@gmail.com}
\authornote{Corresponding author: Yujuan Ding}
\affiliation{
  \institution{The Hong Kong Polytechnic University, HK SAR}}

\author{Liangbo Ning}
\email{BigLemon1123@gmail.com}
\affiliation{
  \institution{The Hong Kong Polytechnic University, HK SAR}}

\author{Shijie Wang}
\email{shijie.wang@connect.polyu.hk}
\affiliation{
  \institution{The Hong Kong Polytechnic University, HK SAR}}

\author{Hengyun Li}
\email{neilhengyun.li@polyu.edu.hk}
\affiliation{
  \institution{The Hong Kong Polytechnic University, HK SAR}}

\author{Dawei Yin}
\email{yindawei@acm.org}
\affiliation{
  \institution{Baidu Inc, China}}
  
\author{Tat-Seng Chua}
\email{dcscts@nus.edu.sg}
\affiliation{
  \institution{National University of Singapore, Singapore}}
  
\author{Qing Li}
\email{csqli@comp.polyu.edu.hk}
\affiliation{
  \institution{The Hong Kong Polytechnic University, HK SAR}}

\begin{abstract}
As one of the most advanced techniques in AI, Retrieval-Augmented Generation (RAG) can offer reliable and up-to-date external knowledge, providing huge convenience for numerous tasks. Particularly in the era of AI-Generated Content (AIGC), the powerful capacity of retrieval in providing additional knowledge enables RAG to assist existing generative AI in producing high-quality outputs.  Recently, Large Language Models (LLMs) have demonstrated revolutionary abilities in language understanding and generation, while still facing inherent limitations, such as hallucinations and out-of-date internal knowledge. Given the powerful abilities of RAG in providing the latest and helpful auxiliary information, Retrieval-Augmented Large Language Models (RA-LLMs) have emerged to harness external and authoritative knowledge bases, rather than solely relying on the model's internal knowledge, to augment the generation quality of LLMs. In this survey, we comprehensively review existing research studies in RA-LLMs, covering three primary technical perspectives: architectures, training strategies, and applications. As the preliminary knowledge, we briefly introduce the foundations and recent advances of LLMs. 
Then, to illustrate the practical significance of RAG for LLMs, we systematically review mainstream relevant work by their architectures, training strategies, and application areas, detailing specifically the challenges of each and the corresponding capabilities of RA-LLMs.  Finally, to deliver deeper insights, we discuss current limitations and several promising directions for future research. Updated information about this survey can be found at \textit{\url{https://advanced-recommender-systems.github.io/RAG-Meets-LLMs/}}\footnote{This is the long version of the survey to appear at KDD 2024~\cite{fan2024rag}}. 
\end{abstract}


\vspace{-0.15in}

\settopmatter{printfolios=true}

\settopmatter{printacmref=false}
\renewcommand\footnotetextcopyrightpermission[1]{}

\maketitle

\input{RAG/Introduction}

\input{RAG/Background}
\input{RAG/Architecture}
\input{RAG/Training}

\input{RAG/Applications}
\input{RAG/Future_work}

\input{RAG/Conclusion}

\bibliographystyle{ACM-Reference-Format}
\balance
\bibliography{bib-refs}

\end{document}

%% file: RAG/Introduction.tex
\section{Introduction}
 

As one of the most fundamental data mining techniques, retrieval 
aims to understand the input query and extract relevant information from external data sources~\cite{kobayashi2000information,singhal2001modern,deng2024deep,ding2020discriminative}. It has 
found extensive application in various fields~\cite{buttcher2016information,yin2016ranking,o2016leveraging,ding2023personalized}, such as search, question answering, and recommender systems.
For instance, search engines (e.g., Google, Bing, and Baidu) are the most successful applications of retrieval in the industry; they can filter and retrieve the most relevant web pages or documents that can match a user's query~\cite{croft2010search,yin2016ranking}, enabling users to find the desired information effectively. 
Meanwhile, retrieval models, through effective data maintenance in external databases, can provide faithful and timely external knowledge, thereby serving vital functions in various knowledge-intensive tasks.
Due to their powerful capacities, retrieval techniques have been successfully incorporated into advanced generative models in the era of AI-Generated Content (AIGC)~\cite{li2023empowering,wu2024coral,sheynin2022knn}.
Notably, the integration of retrieval models with language models has given rise to Retrieval-Augmented Generation (RAG)~\cite{lewis2020retrieval}, which has emerged as one of the most representative techniques in the field of generative AI, 
aiming to enhance the quality of the generated text content with retrieved information~\cite{li2023empowering,lewis2020retrieval,borgeaud2022improving}.

\begin{figure}[t]
\centering
\includegraphics[width=1\columnwidth]{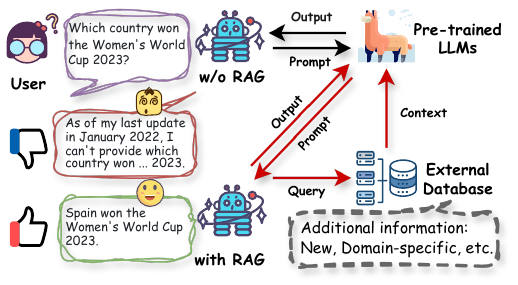}
\caption{ 
Retrieval-Augmented Generation (RAG) meets Large Language Models (LLMs). 
When the user's query is out-of-scope, e.g., unseen content in training data or the need for the latest information for the answer, LLMs might show inferior generation performance. 
With the help of RAG, LLMs can leverage additional relevant information from external database to enhance their text generation capability.
}
\label{fig:task}
\end{figure}

To advance generation models and enhance the generated results, RAG incorporates information or knowledge from external data sources, which serves as supplementary for the input query or the generated output~\cite{min2020ambigqa,khandelwal2019generalization}.
Specifically, RAG first invokes the retriever to search and extract the relevant documents from external databases, which are then leveraged as the context to enhance the generation process~\cite{izacard2021leveraging}.
In practice, RAG techniques are feasible and efficient to apply in various generation tasks with simple adaptation of the retrieval component, requiring minimal or even no additional training~\cite{ram2023context}. Recent studies have demonstrated the great potential of RAG not only for knowledge-intensive tasks such as the Open-domain Question Answering (OpenQA)~\cite{borgeaud2022improving,guu2020retrieval,petroni2020context,shi2024compressing}, but also for general language tasks~\cite{khandelwal2019generalization,he2021fast,xu2020boosting}, and various downstream applications~\cite{wu2024coral,liu2023multi}.

Recent years have witnessed the rapid development of pre-trained foundation models, particularly Large Language Models (LLMs), which have demonstrated impressive performance across various tasks~\cite{chowdhery2023palm,achiam2023gpt}, including recommender systems~\cite{zhao2024recommender}, molecule discovery~\cite{li2023empowering}, and report generation~\cite{ding2024fashionregen}.
Technically, the great success of LLMs can be technically attributed to the advanced architectures with billion-level parameters pre-training on a huge amount of training corpus from various sources. These technical improvements have given rise to the remarkable emergence capabilities of LLMs~\cite{zhao2023survey,zhao2024recommender}, particularly in language understanding and generation, in-context learning, and others. 
For instance, GPT-FAR introduces detailed prompts to teach GPT-4 to perform image tagging, statistical analysis, and text analysis for multi-modal fashion report generation~\cite{ding2024fashionregen}. 
LLMs also achieve promising performance in recommender systems by understanding users' preferences towards items~\cite{wang2024rethinking,zhao2024recommender}. Despite the success, LLMs still suffer from intrinsic limitations~\cite{zhao2024recommender,zhao2023survey}, such as the lack of domain-specific knowledge, the problem of ``hallucination'', and the substantial computational resources required for updating the models.  
These problems are particularly notable in domain-specific fields like medicine and law. 
For instance, a recent study has demonstrated that legal hallucinations are pervasive and disturbing, with hallucination rates ranging from 69\% to 88\% in responses to specific legal queries for state-of-the-art LLMs~\cite{dahl2024large}.
Moreover, the challenges of tackling the hallucination problem become even harder due to the substantial computational resources required for fine-tuning LLMs with domain-specific or the latest data. 
This, in turn, significantly hinders the widespread adoption of LLMs in various real-world applications. 

To address these limitations, recent efforts have been made to take advantage of RAG to enhance the capabilities of LLMs in various tasks~\cite{shi2023replug,khandelwal2019generalization,borgeaud2022improving,izacard2021distilling}, especially those demanding high for the latest and reliable knowledge such as Question Answer (QA), AI4Science, and software engineering. For example, ~\citet{lozano2023clinfo} introduces a scientific QA system based on retrieving scientific literature dynamically. MolReGPT leverages RAG to enhance the in-context learning ability of ChatGPT for molecular discovery~\cite{li2023empowering}. It is also been demonstrated that RAG can effectively reduce hallucinations in conversational tasks~\cite{Shuster2021retrieval,xu2022beyond}. 
As illustrated in Figure~\ref{fig:task}, an LLM-based dialog system will not be able to answer well for out-of-scope queries. With the help of RAG to retrieve relevant knowledge from external
database and integrate it into the process of generation, the dialog system succeeds in giving correct answers. Given the remarkable progress in advancing LLMs with RAG, there is an imperative need for a systematic review of recent advances in Retrieval-Augmented Large Language Models (\textbf{RA-LLMs}).

This survey aims to provide a comprehensive overview of RA-LLMs by summarizing representative methods from the aspects of the architecture, training strategy, and application area respectively. 
More specifically, following a brief introduction to the background knowledge of LLMs in Section~\ref{sec:background}, we review existing research from several primary perspectives of RA-LLMs in terms of retrieval, generation, and augmentation in Section~\ref{sec:architecture}, as well as the necessity and application frequency of retrieval in RAG.
Then, we summarize the main training techniques of RA-LLMs in Section ~\ref{sec:training} and various RA-LLMs applications in Section ~\ref{sec:applications}. 
Finally, in Section~\ref{sec:future}, we discuss key challenges and potential directions for future exploration. 

Concurrent to our survey, several related surveys have diverse focuses for RAG and LLMs. For example, \citet{zhao2023retrieving} specifically review multi-modal information-based RAG techniques and \citet{zhao2024retrieval} discuss the  RAG for AIGC. \citet{gao2023retrieval} conduct a relatively comprehensive overview of RAG for LLMs. Our survey differs from these surveys in concentrating on technical perspectives and systematically reviewing models according to the architecture and training paradigm in RA-LLMs, as well as application tasks.

%% file: RAG/Background.tex
\section{Background}
\label{sec:background}
In this section, we briefly present the background of large language models and prompt learning. 

\subsection{Large Language Models (LLMs)}
Recently, the significant breakthrough of LLMs has revolutionized the field of artificial intelligence~\cite{zhao2023survey,brown2020language,fan2024graph}. 
The advanced LLMs are typically pre-trained on extensive data with billion-level parameters and have demonstrated the ability to understand and generate human-like text, leading to advancements in various natural language processing tasks such as text generation and information retrieval~\cite{zhao2023survey,zhao2024recommender}. 
LLMs can be adapted to a variety of downstream tasks by fine-tuning them on specific datasets, allowing them to specialize in particular domains or applications. 
In general, most existing LLMs can be broadly divided into three main categories: Encoder-only, Decoder-only, and Encoder-Decoder models.
\begin{figure*}[t]
\centering
\includegraphics[width=1\linewidth]{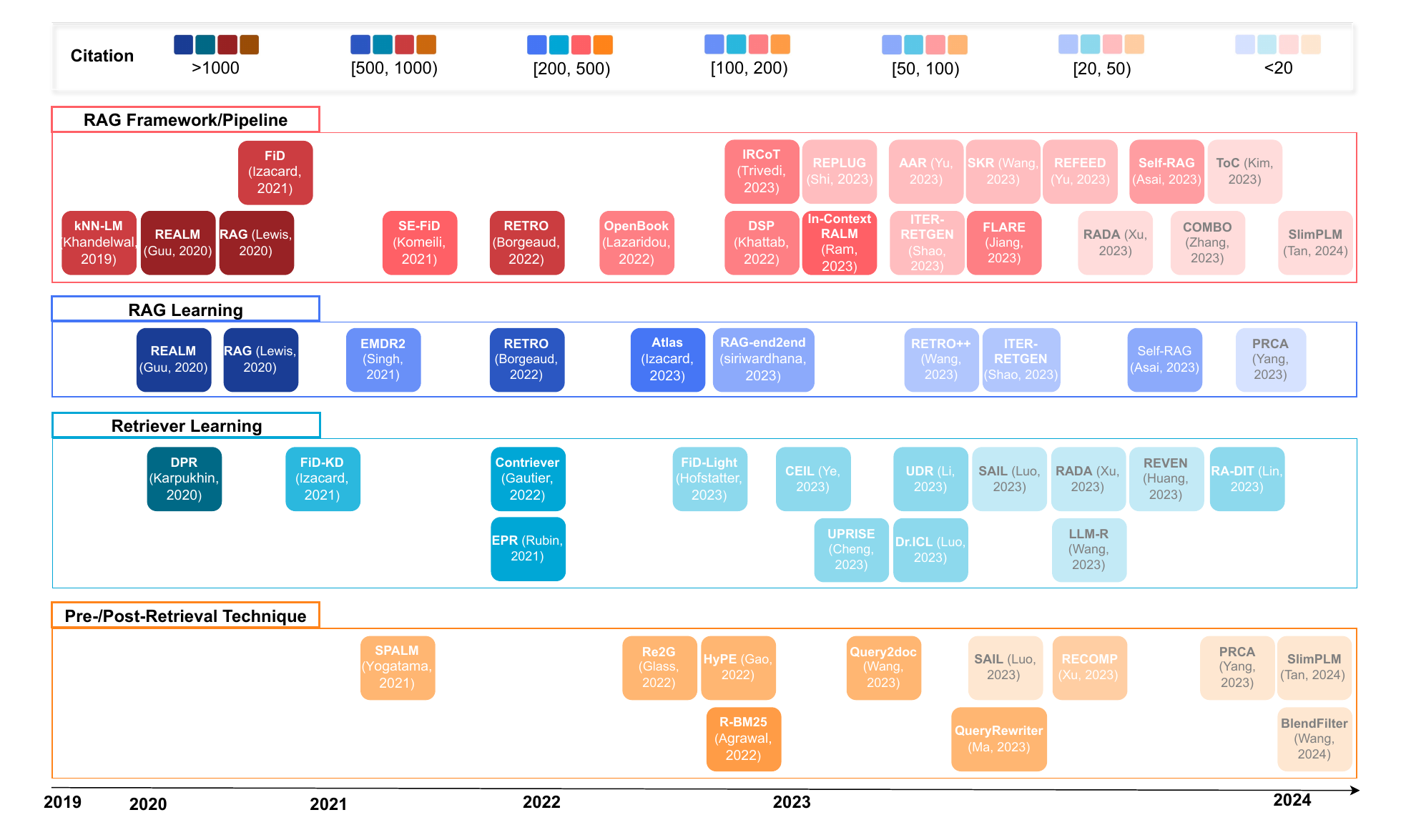}
\caption{
Representing RAG and RA-LLMs methods organized by their main design focus, proposed time and impact (shown by citation). Note that the first author and year shown in the figure along with the model name can be used to locate corresponding reference. 
}
\label{fig:work_org}
\vskip -0.15in
\end{figure*}
Encoder-only models, such as the BERT (Bidirectional Encoder Representations from Transformers)~\cite{devlin2018bert} family of models, process input text by encoding it into a high-dimensional space. 
The key feature of Encoder-only models is their bi-directional nature, meaning that they can take into account both the left and right context of each token when encoding it. This bi-directionality allows Encoder-only models to better understand the meaning of words in context, which is crucial for tasks like sentiment analysis, review reading, and text classification~\cite{xu2019bert,devlin2018bert}.
In contrast to these models, Decoder-only models generate text in a left-to-right fashion.
As a representative Decoder-only model, GPT (Generative Pre-trained Transformer)~\cite{radford2018improving} predicts the next token in a sequence based on the context provided by the previous tokens.
Their architecture makes them particularly effective for tasks like language generation, code generation, and creative writing.
Encoder-Decoder models, such as T5 (Text-To-Text Transfer Transformer)~\cite{raffel2020exploring}, uniquely transform a variety of NLP tasks into text generation problems. 
To be more specific, the encoder in T5 processes the input sequence to capture its meaning, while the decoder generates the output sequence based on the encoded information. This T5 architecture is well-suited for tasks that involve converting one sequence into another, such as machine translation, summarization, and conversational response generation.

\subsection{Prompt Learning}
\subsubsection{Prompting Engineering}
Due to the massive parameters of LLMs, prompt learning emerged as a paradigm to leverage the power of LLM to implement various tasks~\cite{zhao2023survey,zhao2024recommender}, instead of fine-tuning the LLMs extensively.
Prompt learning carefully designs the input that guides the model to perform downstream tasks in LLMs.
For example, early methods~\cite{petroni2019language,brown2020language} provide manually crafted templates to handle various tasks in NLP.
Specifically, Encoder-only models like BERT typically adopt cloze prompts because they very closely match the form of their pre-training task~\cite{petroni2019language,cui2021template}.
For other models like GPT, prefix prompts tend to be more suitable as they mesh well with the generation tasks~\cite{brown2020language}.
However, manually designed prompts rely on human experience without effectiveness guarantees. To address this limitation,  soft prompt tuning was developed to learn the trainable continuous prompt embeddings~\cite{li2021prefix,vu2021spot,tu2022prompt}.
For instance, Prefix-Tuning~\cite{li2021prefix} prepends a series of prefix embedding in the input, which can be trained and updated.
This apportion allows prompts not to be real text, giving more flexibility in the generation of prompts.
However, due to the lack of domain-specific knowledge, the model might still not generate accurate responses when facing new tasks.

\subsubsection{In-Context Learning (ICL)}
To overcome the limitations of vanilla prompt learning, recent efforts~\cite{liu2021makes,kim2022self,zhang2022automatic} have developed in-context learning (ICL).
ICL is a specific method of prompt learning that gives the model a few demonstrations of tasks within the prompt.
This paradigm allows pre-trained LLMs to understand the pattern provided by the demonstrations to solve novel tasks without the need for fine-tuning. 
For example, by carefully selecting a few demonstrations, GPT-3~\cite{brown2020language} has shown the capability to perform few-shot tasks~\cite{liu2021makes}. 
This success indicates that LLMs have a remarkable ability to rapidly adapt to new tasks based on task-specific knowledge.

Despite its effectiveness, ICL usually relies heavily on the quality of the provided demonstrations~\cite{liu2022makes,Su2023selective}, which may lead to the generation of sub-optimal outputs.
Even worse, ICL may not have enough necessary information or prior knowledge to guide the LLMs in generating accurate responses. 
To address the aforementioned limitations of ICL, more recent studies introduce Retrieval-Augmented Generation (RAG) technologies~\cite{lewis2020retrieval,
ram2023context,shi2023replug}. 
By integrating retrieval with generation, RAG models provide a promising direction for enhancing the performance and adaptability of LLMs across various tasks.

%% file: RAG/Architecture.tex
\section{Retrieval-Augmented Large Language Models (RA-LLMs)}
\label{sec:architecture}

The RAG framework in the era of LLMs consists of several major processes: \emph{retrieval}, \emph{generation}, and \emph{augmentation}, as well as the mechanism to determine whether the retrieval is needed. In this section, we will introduce important techniques involved in each process. 

\begin{figure*}[t]
\centering
\includegraphics[width=1\linewidth]{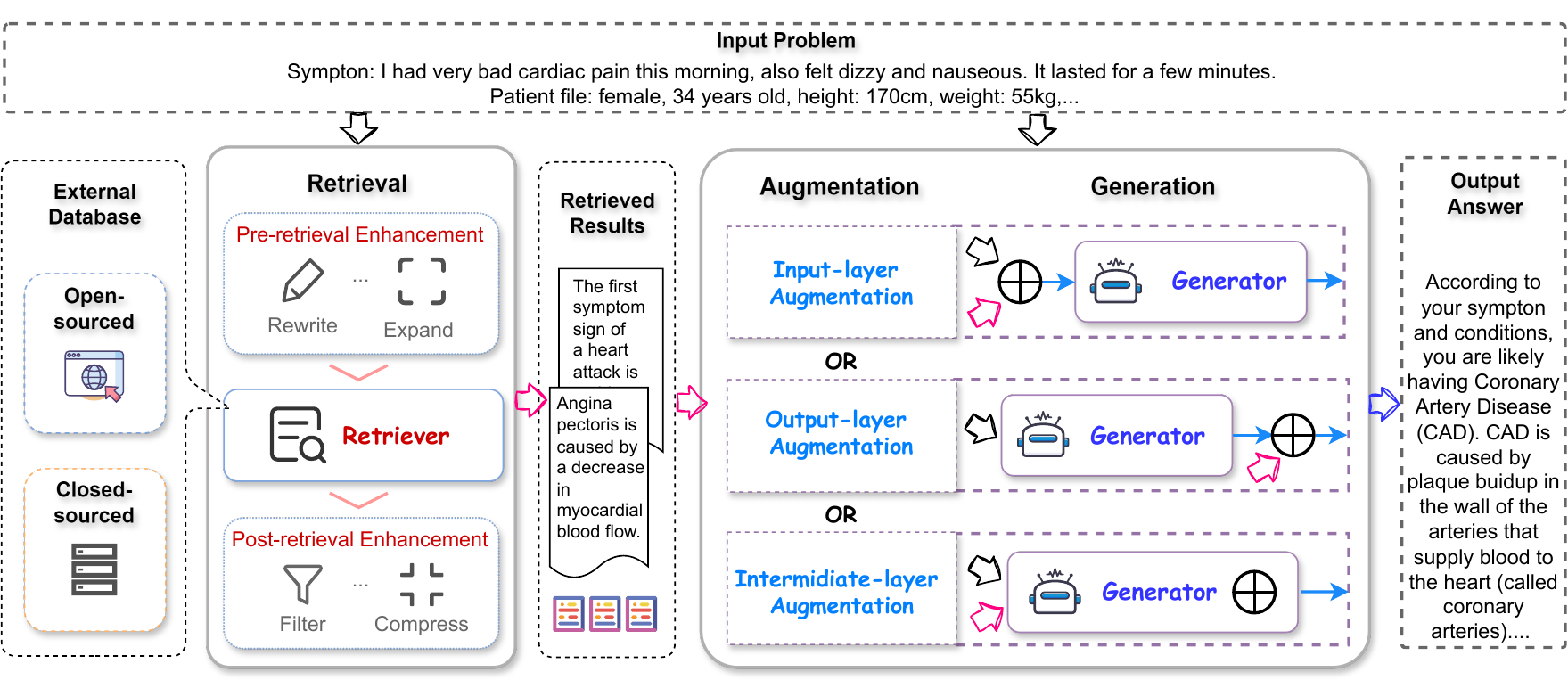}
\caption{
Illustration of the basic Retrieval-Augmented Large Language Models (RA-LLMs) framework for a specific QA task, which consists of three main components: retrieval, augmentation, and generation. Retrieval may have different procedures with various designs, which optionally includes pre-retrieval and post-retrieval processes. The retrieved documents are further leveraged in generation with the augmentation module, which may be at different integration stages.
}
\label{fig:arc}
\end{figure*}

\begin{figure}[t]
\centering
\includegraphics[width=0.99\columnwidth]{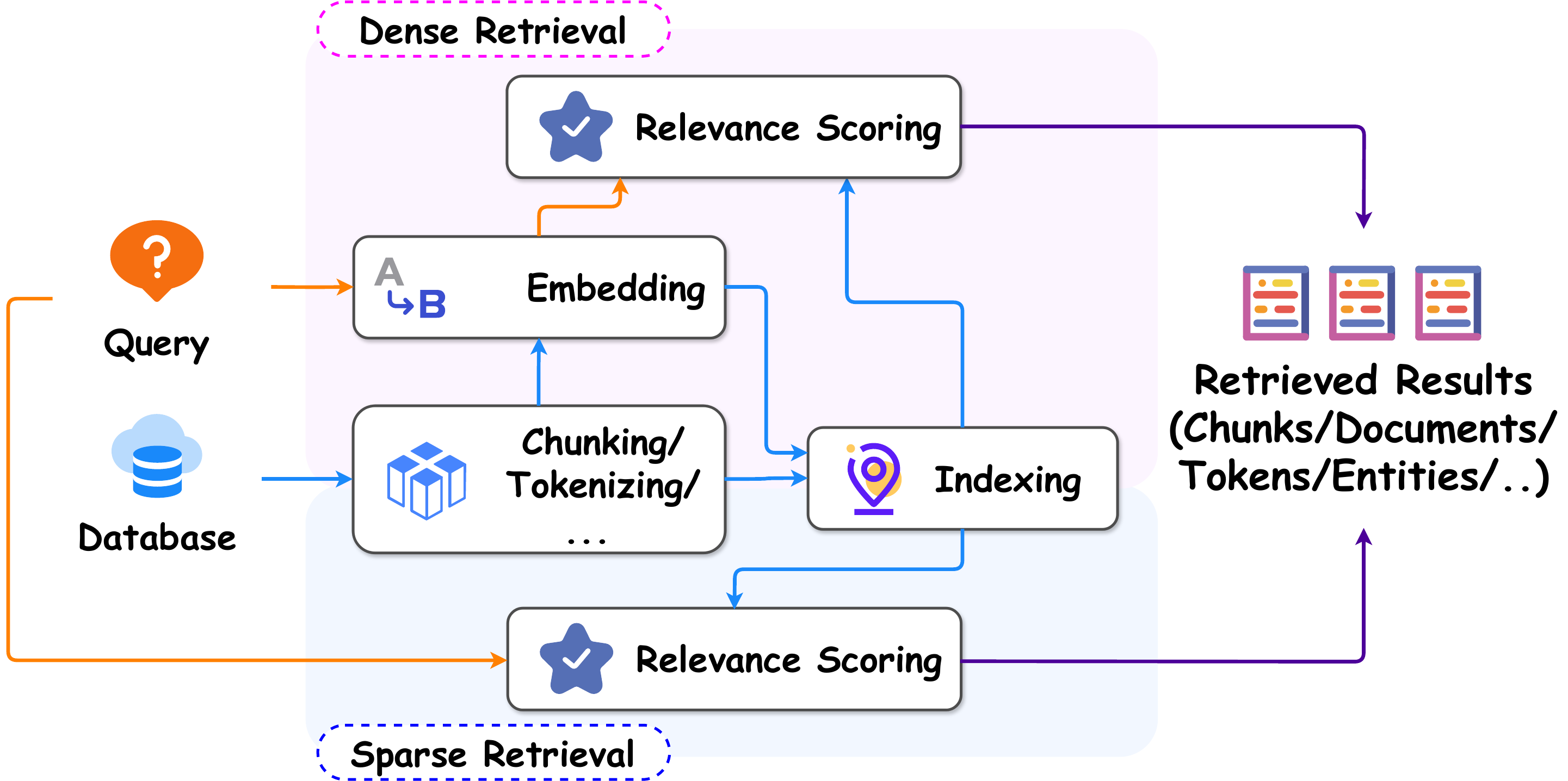}
\caption{Illustration of the retriever in RA-LLMs, which can be implemented in either dense or sparse manners, each with several key operations. 
}
\label{fig:retriever}
\end{figure}

\subsection{Retrieval} 
Given the query from the input of LLMs, the retrieval process in RAG aims to provide relevant information from the external knowledge sources, which can be either open-sourced or closed-sourced as shown in Figure~\ref{fig:arc}. The key component, retriever, as further detailed in Figure~\ref{fig:retriever}, consists of several procedures, functioning as a whole to measure the relevance between the query and documents in the database for effective information retrieval. The specific pipeline of the retrieval is further determined by whether the pre- and post-retrieval processes are included. 
In this subsection, we will introduce the major techniques involved in the retrieval of traditional and LLM-based RAGs, including the retriever type, retrieval granularity, pre- and post-retrieval enhancement, and database construction. 

\subsubsection{Retriever Type}
Retrieval methods can be generally categorized into two types: sparse and dense, based on the information encoding methods. Sparse retrieval is word-based and applied in text retrieval mostly, while dense retrieval embeds queries and external knowledge into vector spaces and can applied to various data formats.  

As a straightforward approach, sparse retrieval, e.g.,  TF-IDF and BM25~\cite{sparck1972statistical, robertson2009probabilistic}, usually relies on inverted index matching along with the raw data input. For example, many studies directly apply BM25 for passage-level retrieval to facilitate their RAG~\cite{chen2017reading,ram2023context,zhong2022training,jiang2023active,zhou2022docprompting,xu2023recomp}, where passages are specifically represented as a bag of words and ranked based on term and inverse document frequencies~\cite{izacard2021leveraging}. On top of offering supplementary to enhance the input of the generator, sparse retrieval has also been used to find demonstrations to function in in-context learning for RA-LLMs~\cite{ye2023compositional,luo2023dr,rubin2021learning, agrawal2022context, sia2023context}. 
The main limitation of applying sparse retrieval in RAG is its no-training nature, which makes the retrieval performance heavily rely on the quality of the database and the query. Moreover, such fixed term-based methods only support similarity-based retrieval, while cannot be adapted for other retrieval criteria possibly existing in LLM applications, such as the diversity~\cite{drozdov2022compositional}. 

Dense retrieval, on the contrary, embeds the query and documents into continuous vector space with certain criteria, for example, semantic similarity~\cite{karpukhin2020dense}. Dense retrieval methods are usually trainable, therefore holding more flexibility and potential in adaptation. As the key component of dense retriever, the embedding models have delicately different designs in existing RAG models. A simple design~\cite{khandelwal2019generalization,lewis2020pre,wu2022memorizing} is to directly use a part of the generation model as the embedding layer of the retriever, which might be able to enhance the alignment between the retrieval and generation processes. BERT-based backbone~\cite{devlin2018bert} is widely applied in retrieval models. One common retriever design in RAG is to construct two-stream encoders with the BERT structure (one encoder for the query and the other for the documents), which is also called bi-encoder~\cite{wu2019scalable,shi2023replug}.
Early-stage RAG methods tend to freeze~\cite{borgeaud2022improving,ram2023context} or partially freeze~\cite{lewis2020retrieval} the parameters of the retriever to perform general-level relevant knowledge extraction and pay more attention to the knowledge leveraging and generator fine-tuning. Large-scale specialized pre-training further enhances RAG models to excel in more knowledge-intensive tasks. One typical success is Dense Passage Retriever (DPR)~\cite{karpukhin2020dense}, which uses a BERT-based backbone and is pre-trained specifically for the OpenQA task with question-answer pair data. DPR has shown strong capacity as a pre-trained retriever, facilitating many RAG models to succeed in various downstream tasks~\cite{lewis2020retrieval,izacard2021leveraging,siriwardhana2023improving,singh2021end,shi2023replug}. It has also been regarded as the first step in the RAG paradigm for improving the performance of LLMs, which may further enhance the alignment of the embeddings between queries and relevant textual data through fine-tuning~\cite{cheng2023uprise}. A recent study~\cite{benjamin2024} has also discovered that DPR training decentralizes how knowledge is stored in the network, creating multiple access pathways to the same information. With effective fine-tuning, bi-encoder retrievers are also applied widely in ICL-based RAG~\cite{rubin2021learning,poesia2022synchromesh,lu2022dynamic,ye2023compositional,milios2023context,li2023mot}. Specifically, they have been more often used for sentence embedding similarity-based retrieval, as well as for some special requirement in ICL, such as diverse example retrieval~\cite{ye2023compositional}.

Another stream of dense retrievers having been widely applied in RA-LLMs uses one encoder only, which may be based on Transformer, BERT or other off-the-shelf sequence modeling backbones. These one-encoder retrievers are generally pre-trained on large-scale unaligned documents by contrastive learning~\cite{benjamin2024}, which may therefore excel for their versatility, meaning that they can transfer and generalize better to new domains or tasks. Such general-purpose pre-trained retrievers, 
e.g.,  Contriever~\cite{gautier2022unsupervised} and Spider~\cite{ram2021learning}, would be more flexible to use in LLMs targeting on various tasks and have demonstrated their effectiveness in many RA-LLM methods, such as In-Context RALM~\cite{ram2023context}, Atlas~\cite{izacard2023atlas} and Self-RAG~\cite{asai2023self}.
According to experimental results in existing studies~\cite{yu2023generate}, for open-domain QA tasks, when cooperated with InstructGPT~\cite{ouyang2022training}, applying general-purpose pre-trained retriever (Contriever) without fine-tuning achieves comparable performance to sparse retriever (BM25). However, they are both worse than the DPR model fine-tuned on target datasets, showing the effectiveness of fine-tuning on targeted tasks and data. 

\subsubsection{Retrieval Granularity}
Retrieval granularity denotes the retrieval unit in which the corpus is indexed, e.g., document, passage, token, or other levels like entity. For RAG, the choice of retrieval granularity can significantly impact the overall performance of the model in terms of effectiveness and efficiency as they determine the saving space for the database as well as the computational cost for searching~\cite{asai2023retrieval}. Early stage retrieval-augmented language models~\cite{chen2017reading} propose to retrieve whole pieces of documents, and then apply a machine comprehension model trained to detect answer spans in the returned documents, which focuses more on language reading and key information locating in the document. 
In generative language models, \textbf{Chunk retrieval} (also called passages in some references~\cite{karpukhin2020dense,guu2020retrieval,jiang2023active}) is common, which has been used in both traditional and LLM-based RAG models such as REALM~\cite{guu2020retrieval}, RAG~\cite{lewis2020retrieval} and Atlas~\cite{izacard2023atlas}. A more fine-grained retrieval, i.e., \textbf{token retrieval}, instead can be done with faster searching but will bring more burden for the database saving. Token retrieval is more suitable in cases requiring rare patterns or out-of-domain data~\cite{khandelwal2019generalization}, meanwhile cooperates well with the every-token retrieval strategy as applied in kNN-LM and other similar work~\cite{yogatama2021adaptive,he2021efficient,min2022nonparametric}. In comparison, a text chunk may contain compact and complete information with less redundancy and irrelevancy, therefore becoming the mainstream retrieval text granularity in RAG. 

Another major retrieval granularity proposed in RAG is \textbf{entity retrieval}. Unlike the above types of granularity, entity retrieval is designed from the perspective of knowledge rather than language. \citet{fevry2020entities} introduce the Entities as Experts (E{\small A}E) model, which divides the parameter space of language models according to the entity identity. The proposed E{\small A}E model aims to learn entity representations from the text along with other model parameters with the Wikipedia database and represent knowledge with entity memory.
At a more fine-grained level, \citet{de2021mention} propose to build the knowledge base by learning and retrieving mention rather than entity. Overall, applying entity or mention-level retrieval in RAG would be more effective for entity-centric tasks, and more efficient in space compared to token-wise retrieval.

\subsubsection{Pre-retrieval and Post-retrieval Enhancement}
To ensure the retrieval quality, i.e., increase the accuracy and relevance of the retrieved results, various pre- and post-retrieval strategies have been proposed to further enhance the input and output of the retriever.  \citet{wang2023query2doc} propose a \textbf{query expansion} approach Query2doc, which generates pseudo-documents by few-shot prompting LLMs and expands the query with the relevant information in pseudo-documents to improve the query disambiguation and guide the retrievers. They have empirically demonstrated that such a method can boost the performance of both the sparse and dense retriever~\cite{karpukhin2020dense} on ad-hoc information retrieval datasets. Similarly, \citet{gao2022precise} propose Hypothetical Document Embedding (HyDE) method, which instructs an LLM to generate hypothetical documents for the given query. The hypothetical documents are then used as new queries to get embedded and search for neighbors with the dense retriever. 

Another pre-retrieval strategy, \textbf{query rewrite}~\cite{ma2023query}, aims to close the gaps between the input text and the needed knowledge in retrieval, to reformulate the original question into a more conducive version to retrieve. Specifically,  \citet{ma2023query} propose the Rewrite-Retrieve-Read framework, which prompts an LLM to generate the query for the retrieval function. The motivation of the rewriting step is to clarify the retrieval need in the new query to ease the burden on the retrieval function to comprehend the input and enhance the output, i.e., retrieved relevant information. They have tested both the settings of using a frozen LLM and a trainable model to be the rewriter, both outperforming naive RAG or generation models, demonstrating diverse performance on different tested QA datasets though. \citet{tan2024small} also formulate a query rewriting strategy in their model that decomposes the heuristic answer from a proxy generation model into distinct claims. 

\citet{yu2023improving} propose \textbf{query augmentation} to combine the original query and the preliminary generated outputs as a new query, which is further used to retrieve relevant information from the external database. The retrieved results can inspire the language model to rethink the generated results and enhance them. Compared to applying only the original query, such augmentation may contribute more relevant information retrieved from the corpus for the directly clarification of query-output relationships. Including initial output in the new query further enhances the lexical and semantic overlap between the supporting documents to be retrieved with the given question. Query augmentation achieves overall better performance among these query enhancement strategies since it may process all retrieved knowledge collectively while generating answers~\cite{wang2024blendfilter}.

Post-retrieval enhancement denotes the procedure to process  the  extracted top-k documents from the retriever before feeding them to the generator for the sake of better alignment between the retrieval and generation stages~\cite{yang2023prca}, particularly for closed-source generators such as LLMs. For example, \citet{yang2023prca} propose the Pluggable Reward-driven Context Adapter (PRCA) that enables to fine-tune the lightweight adapter instead of  the generator on specific datasets. It also distills the retrieved documents through reinforcement learning  with the rewards resulting from  the generator. \citet{glass2022re2g} propose Retrieve-Rerank-Generate (R$^2$G) method, which assembles the retrieved documents of different retrieval approaches with the rerank operation to boost the robustness of the retrieval results. Another consideration for applying post-retrieval enhancement is that  the retrieved information may sometimes be irrelevant or contain noise, which might not help with the generation model for the task, or even worse, harm the generation process~\cite{wang2023self}. \citet{wang2023self}, \citet{asai2023self}, \citet{yu2023improving} propose different strategies to mitigate the noise in retrieved knowledge documents. However, \citet{xiong2023can} empirically studied that these methods are dependent on the LLM's confidence levels, which might not be as precise as expected. For this problem, \citet{wang2024blendfilter} propose BlendFilter, which simultaneously considers the pre-retrieval query generation blending and the post-retrieval knowledge filtering. This method can tackle the complex questions as well as  the noisy retrieved knowledge problems, therefore comprehensively enhancing the RA-LLM performance. 

More recently, advanced RAG pipelines have been proposed using LLMs to generate reasoning paths and plans with the Information Retrieval (IR) module to iteratively retrieve knowledge to enhance LLM-based generation~\cite{yao2022react,xu2023search,shao2023enhancing}. However, \citet{zhu2023furthest} point out that if the outputs of IR and LLM are low-quality, the retrieval and generation processes will get hindered by each other with such an iterative guidance pipeline. To overcome this barrier, they propose a new reasoning approach for query and retrieved knowledge enhancement. Post-retrieval strategies may also function to enhance the compatibility between the retrieved results and the generation models. For example, one of the main limitations of existing LLMs is the length of the input tokens,  which prevents long retrieved documents being directly incorporated into existing RA-LLMs. For this limitation, \citet{xu2023recomp} propose Retrieve, Compress, Prepend (RECOMP), which adds an intermediate step to process the retrieved documents into a textual summary before in-context augmentation in the generation process. From another perspective, long retrieved passage list leads to a high inference latency when using auto-regressive decoding at generation stage, which hurts the model's efficiency. For this limitation, \citet{hofstatter2023fidlight} propose a light version of FiD model that compresses the encoded vectors per retrieved passage before concatenating and feeding them through the decoder and also includes a re-ranker on the retrieved results before applying them in the generation. 

\begin{table*}[h]
    \centering
    \begin{tabular}{p{6mm}p{18mm}p{6mm}p{16mm}p{10mm}p{10mm}p{14mm}p{20mm}p{8mm}p{30mm}}
    \hline
    Time &Model & Cite & Retriever & RetTrain &RetAug Stage & Pre-/Post-Retrieval &Generator &Aug &Evaluation\\ \hline
    2019 &kNN-LM~\cite{khandelwal2019generalization} &619 &DR(GP) &No &Inf &RA &DT &Output &LG\\
    2020 &REALM~\cite{guu2020retrieval} &1437 &DR(BE,BT) &Yes &PT+FT &/ &ET &Input &OpenQA(NQ, WQ, CT) \\
    2020 &RAG~\cite{lewis2020retrieval} &2125 &DR(DPR) &Yes &FT &/ &ED (BART) &Input &OpenQA, AQA, Jeopardy QG, FV \\
    2021 &FiD~\cite{izacard2021leveraging} &780 &SR(BM25)/ DR(DPR) &No &FT &/ &ED (T5/BART) &Input &OpenQA \\
    2021 &SE-FiD~\cite{komeili2021internet} &286 &SE(Bing) &No &Inf &RQG &FiD &Input &WizInt, WoW\\
    2021 &FiD-KD~\cite{izacard2021distilling} &190 &DR(BE) &Yes &FT &CR &FiD &Input &OpenQA\\
    2021 &RETRO~\cite{borgeaud2022improving} &683 &DR(BERT, DPR) &No &PT &/ &ED &Inter &LM, OpenQA \\
    2021 &EPR~\cite{rubin2021learning} &384 &DR(DPR) &Yes &Inf &CR &GPT-3,J,Neo, CODEX &Demon &UR\\
    2022 &OpenBook~\cite{lazaridou2022internet} &145 &SE+SR &No & &QE &GOPHER LM &Input &QA, FV \\
    2022 &DSP~\cite{khattab2022demonstrate} &117 &ColBERTv2 &No &Inf &RQG, RF  & GPT-3.5 &Demon  &OpenQA, MHQA, CQA\\
    2023 &In-Context RALM~\cite{ram2023context} &211 &DR/SR &No &Inf &TRR &GPT-2,J,Neo &Input &LM, OpenQA \\
    2023 &Atlas~\cite{izacard2023atlas} &367 &DR(OE) &Yes &PT+FT &/ &ED &Input &OpenQA, FV, WoW, EL,SF, MMLU \\
    2023 &FLARE~\cite{jiang2023active} &133 &SR(BM25)/ SE(Bing)  &No &Inf &RQG &GPT-3.5 &Input &MHQA, CR, LongQA, OS\\
    2023 &IRCoT~\cite{trivedi2023interleaving} &114 &SR(BM25) &No &Inf &/ &GPT-3,Flan-T5 &Input&OpenQA\\
    2023 &Self-RAG~\cite{asai2023self} &85 &DR(OE) &No &FT &CM  & tunable LLM  & &OpenQA, LongQA, FV, BG\\
    2023 &REPLUG~\cite{shi2023replug} &48 &DR(BE) &Yes &FT &TRA  &GPT-2,3 &Inpput &MMLU, OpenQA\\
    2023 &UDR~\cite{Li2023unified} &42 &DR(DPR) &Yes &FT &CR &GPT-Neo &Demon &40 NLP tasks\\
    2023  &ITER-RETGEN~\cite{shao2023enhancing} &40 &DR(DPR) &Yes &FT &RR &InstructGPT, Llama-2 &Input &MHQA, FV, CR  \\
    \hline
    \end{tabular}
    \caption{Basic publication information and main technical designs of high-impact RAG and RA-LLM models.$^1$}
    \label{tab:model_org}
    \end{table*}
    \footnotetext{Retrievers:
    [BE: Bi-Encoder (also referred as dual encoder), 
    OE: One-Encoder, 
    BT: BERT-style Transformer, 
    GP: Partial Generation, 
    SE: Search Engine, 
    SR: Sparse Retrieval, 
    DPR:\cite{karpukhin2020dense}],
    Generators:
    [DT: Decoder-only Transformer, 
    ET: Encoder-only Transformer, 
    ED: Encoder-Decoder], 
    Pre-/Post-Retrieval techniques:
    [RQG: Retrieval Query Generation, 
    QE: Query Expansion, 
    (T)RR: (Trainable) Re-Ranker,
    TRA: Trainable Retriever Adaptor, 
    CR: Candidate Retrieval, 
    CM: Critic Model],
    Augmentations:[Output: Output-layer Integration, 
    Input: Input-layer Integration, 
    Inter: Intermediate-layer Integration, 
    Demon: As demonstration], 
    Tasks:
    [AQA: Abstractive Question Answering, 
    QG: Question Generation, 
    NQ: Natural Questions, 
    WQ: WebQuestions, 
    CT: CuratedTrec, 
    FV: Factor Verification, 
    TQA: TriviaQA, 
    WizInt: Wizard of the Internet task, 
    WoW: Wizard of Wikipedia task, 
    MHQA: Multi-hop QA, 
    CQA: Conversational QA, 
    EL: Entity Linking, 
    SF: Slot-filling, 
    MMLU: Massively-Multitask Language Understanding, 
    CR: Commonsense Reasoning, 
    LongQA: Long-form QA, 
    OS: Open-domain Summarization, 
    BG: Biography Generation, 
    UR: Utterance Representing, 
    RF: Retrieval Fusion]}

\subsubsection{Database}
Retrieval in RAG is conducted based on external knowledge source, which can be a closed- or open-sourced~\cite{ma2023query,menick2022teaching}, as illustrated in Figure~\ref{fig:arc}. Closed-sourced database generally stores key-value pairs for knowledge, which can be constructed in various ways. The keys are primarily used for similarity matching, being as sparse vectors such as in BM25 or dense embeddings from the retrieval encoding. The value depends on the specific retrieval target, which is raw text in most cases~\cite{guu2020retrieval,lewis2020retrieval,izacard2021leveraging,borgeaud2022improving,lewis2020pre,seo2019real}. 
For example, each Wikipedia article is split into disjoint 100-word chunks, to make a total of 21M documents in  early RAG~\cite{lewis2020retrieval}. Each document is encoded by a dense embedding and saved in the database as the value and key, respectively. The value can store tokens too, one for each as applied in kNN-LM~\cite{khandelwal2019generalization} and SPALM~\cite{yogatama2021adaptive}. The source of the database depends on the specific application domains and tasks. Wikipedia is one of the most commonly applied general retrieval sets in previous RAG work, which stores factual structured information and has several versions differing in scale, from billion token-level~\cite{khandelwal2019generalization, yogatama2021adaptive,lewis2020retrieval,guu2020retrieval,fevry2020entities,de2021mention,xu2023recomp,shi2023replug,ram2023context} to trillion token-level~\cite{borgeaud2022improving}. 
Domain-specific database is also used for downstream tasks. For example, for the code generation task, \citet{zan2022language} collect API information and code files of public libraries to build their APIretriever database. In addition, Zhou et al~\cite{zhou2022docprompting} propose to use a documentation pool frequently updated with new content (newly released libraries) in their model.

Applying Internet searching engine~\cite{luo2023sail} such as Bing and Google avoids the maintenance of the search index and can access up-to-date knowledge~\cite{komeili2021internet,lazaridou2022internet}. Meanwhile, it provides a broader knowledge base than the closed-sourced database~\cite{asai2023self,lazaridou2022internet}. It can also provide high-quality ranking after being tuned over decades of use. Internet search has been widely incorporated with black-box LLMs and shows effectiveness for different functions such as knowledge augmentation~\cite{lazaridou2022internet}, fact-checking~\cite{menick2022teaching} and LLM agent enhancement~\cite{yao2022react}. Compared to traditional RAG, Internet search has been leveraged more as the retriever in RA-LLMs owing to the extraordinary capability of LLMs to be the Reader to comprehend the searching results, i.e., the retrieved documents, as well as LLMs' ability to use tools to process and analyze the them~\cite{ma2023query}. Existing studies~\cite{yu2023generate} have shown that leveraging search engines (e.g., InstrucGPT) is particularly effective for LLMs on zero-shot knowledge-intensive tasks such as OpenQA and fact checking.

\subsection{Generation}
The design of the generator heavily depends on the downstream tasks. For most text generation tasks, Decoder-only and Encoder-Decoder are two dominant structures~\cite{zhao2023survey}. 
The recent development of commercial closed-sourced large foundation models makes black-box generation models mainstream in RA-LLMs. 
In this part, we will briefly review studies with these two types of generators: parameter-accessible (white-box) and parameter-inaccessible (black-box). 

\subsubsection{Parameter-Accessible Generators (White-box)}
The structure of \emph{Encoder-Decoder} processes the input and the target independently with different sets of parameters, in which a cross-attention component is developed to connect input tokens to target tokens. Representative Encoder-Decoder models include T5~\cite{raffel2020exploring} and BART~\cite{lewis2019bart}.
In comparison, \emph{Decoder-only models} process inputs and targets after concatenation, which makes the representations of the two parts concurrently built layer-by-layer as they propagate up the network. 
These two types of generators are widely applied in existing RAG work. 
For example, RAG~\cite{lewis2020retrieval} and Re$^2$G~\cite{glass2022re2g} employ BART; 
FID~\cite{izacard2021leveraging} and EMDR$^2$ utilize T5. There are other models~\cite{borgeaud2022improving,li2022decoupled} leveraging Transformer-based Encoder-Decoder architecture but with some customized design. Generators in RAG differ themselves from general ones by incorporating retrieved data to enhance the generation accuracy and relevance. Furthermore, white-box generators allow parameter optimization, which can be trained to adapt to different retrieval and augmentation approaches for a better performance of generation.

\subsubsection{Parameter-Inaccessible Generators (Black-box)}
A certain proportion of LLMs are released without the disclosure of internal structures or the accessibility of parameters, especially those particularly large-scale ones such as GPT series~\cite{achiam2023gpt}, Codex~\cite{chen2021evaluating} and Claude, which are called black-box generation models. 
These generators only allow the operations of feeding queries (input) and receiving responses (output) while not allowing the internal structure to be altered or parameters to be updated. 
From another perspective, LLMs, even those open for fine-tuning, are large in scale and difficult to tune for downstream domain-specific tasks with only a limited amount of data. 
Black-box RA-LLMs, therefore, focus more on the retrieval and augmentation processes, trying to enhance the generator
by augmenting the input (also called prompt in the context of LLMs)  with  better knowledge, guidance, or examples for the generation. 
For example, \citet{rubin2021learning} proposes to train a prompt retriever with the data labeled by language models themselves, which can be used to provide better examples for in-context learning, therefore enhancing the final generation performance. 
\citet{xu2023recomp} propose to compress the retrieved documents before in-context integration, which can reduce the computational costs and also relieve the burden of LMs to identify relevant information in long retrieved documents.

\subsection{Retrieval Integration for Generation Augmentation}
Augmentation describes the technical process that integrates retrieval and generation parts, which is the essential part of RA-LLMs. In this subsection, we introduce three main designs of augmentation, which are conducted at the input, output, and intermediate layers of generator respectively, as illustrated in Figure~\ref{fig:arc}.

\subsubsection{Input-Layer Integration}
A common way to integrate retrieved information/documents is to combine them with the original input/query and jointly pass them to the generator, which is called input-layer integration. 
For example, In-Context RALM~\cite{ram2023context} applies input-layer integration by specifically concatenating the original input and all retrieved documents into a single sequence as the new input for the generation model. 
Despite the effectiveness, such integration is limited to the number of retrieved documents, since the concatenated new input may be too long to be processed by the generation model. In-context RALM specifically alleviates this limitation by removing tokens from the beginning of the new input. 
To avoid information loss with such a token removing strategy, FID~\cite{izacard2021leveraging} employs a different integration method that processes each retrieved document independently in the encoder. This strategy is scalable to a large number of contexts as it only performs self-attention over one context at a time in the follow-up processing. Atlas~\cite{izacard2023atlas} and R{\small E}P{\small LUG}~\cite{shi2023replug} apply a similar parallel integration by concatenating the query and one retrieved document at a time.
In general, most black-box generation-based RAG methods apply input-layer integration since neither the intermediate layer of the generation model or the output distribution is accessible. 

More specially for LLMs, input-layer integration may use the retrieved content as (additional) prompts or demonstrations on top of using it as supplementary to the original input as in traditional RAGs~\cite{rubin2021learning}. Prompt retrieval aims to find suitable natural language prompts automatically through retrieval to teach the LLM to learn in context~\cite{brown2020language} or to induce the LLM to reason\cite{wei2022chain}. It may boost the zero-shot ability of LLMs without delicate prompt engineering. For example, \citet{cheng2023uprise} propose to learn a prompt retriever based on the input-prompt pair data with score labels resulting from a frozen LLM.

\subsubsection{Output-Layer Integration}
Another kind of augmentation is post-hoc, i.e., output-layer integration, which joints retrieval and generation results. 
For example, kNN-LM~\cite{khandelwal2019generalization} interpolates two next-token distributions in prediction: one induced by the LM and the other induced by the nearest neighbors from the retrieval corpus. 
Output-layer linear integration~\cite{grave2016improving,zhong2022training} is flexible to apply since it can be plugged into most generation models without additional training. However, the simplicity of output-layer integration also limits the model's ability to reason about the retrieved text. To tackle this limitation, \citet{yogatama2021adaptive} propose to add an extra gating network to post-process the retrieved data and achieve comparatively better performance. For LLMs, output-layer integration is as reasonable  and adaptive as  input-layer integration. R{\small E}F{\small{EED}}~\cite{yu2023improving} proposes an answer refining mechanism that applies an LLM to evaluate the retrieved information and adjust the initial answer accordingly to enhance the accuracy of the response. Similarly, \citet{zhang2023merging} propose the COMBO  framework, which matches LLM-generated passages with retrieved counterparts into compatible pairs based on pre-trained discriminators. The passage pairs are then handled by a Fusion-in-Decoder-based~\cite{izacard2021leveraging} to derive a final answer. 

\subsubsection{Intermediate-Layer Integration}
Compared to the above two non-parametric approaches, a more engaging augmentation is to design a semi-parametric module to integrate the retrieved results through the internal layers of the generation model, which is called intermediate-layer integration. 
Such integration might add additional complexity and is promising to enhance the capability of the generation model with effective training.
Typically, a Transformer module is introduced to leverage retrieved information (mostly encoded into dense representations) into the generation model to interact with the representations in the middle stage of the generation. 
For example, R{\small ETRO}~\cite{borgeaud2022improving} introduces a Chunked Cross Attention (CCA) layer to process the retrieved chunks in the generator blocks, and ~\citet{wu2022memorizing} introduces the kNN-Augmented Attention Layer.
Similarly, E{\small A}E~\cite{fevry2020entities} and {\small TOME}~\cite{de2021mention} use Entity Memory and MemoryAttention layer to incorporate the retrieved Entity and Entity Mentions, respectively. 
Such intermediate-layer integration can use many blocks frequently and efficiently to enhance the capability of the whole RAG model. It offers an efficient alternative to incorporate a large number of text chunks frequently retrieved, which are challenging to process with input-layer integration due to the input length limit  of LMs~\cite{borgeaud2022improving}.  However, it also needs to be noted that intermediate-layer integration requires high access to the generation models, which is not  feasible  for most LLMs that are accessible through inference APIs~\cite{ma2023query}. 

\subsection{Retrieval Augmentation Necessity and Frequency}
The retrieval operation in LLM-based generation generally aims to supplement knowledge to enhance generation. 
Although retrieval-augmented models have emerged promising, they have been criticized for not being a universal solution~\cite{li2022large,petroni2020context} as indiscriminately augmenting LLMs with irrelevant passages can override potentially correct knowledge already possessed by LLMs and result in incorrect responses instead~\cite{maekawa2024retrieval}.  
\citet{thakur2023nomiracl} contribute a human-annotated dataset to help evaluate the robustness of LLMs against errors in external retrieved knowledge and observe that LLMs may double the hallucination rate on the non-relevant retrieved passages than on the relevant ones. 
Therefore, it is critical for RA-LLMs to accurately recall the prior knowledge while selectively incorporating retrieved information  only when necessary, 
which is the path to robust RA-LLMs.

Most existing methods determine the necessity of retrieval based on the preliminary answers of LLMs or their internal reasoning results~\cite{ram2023context,min2022rethinking}. For example, Self-RAG~\cite{asai2023self} introduces special tokens to assess the necessity of retrieval and control retrieval behavior. Other methods design iterative prompts to decide if extra information is required during generation, which thereby needs to invoke retrieval or other actions for LLMs~\cite{yao2022react,wei2022chain}. \citet{wang2023self} propose Self-Knowledge guided Retrieval augmentation (SKR) method, which uses LLMs themselves or explicit small trainable models to offer self-knowledge as the reference for the adaptive calling of a retriever.  In traditional RAGs, retrieval necessity judgment has also been explored and proposed to address by intuitive approaches such as assessing the confidence of the logits produced by the generation models~\cite{jiang2021can,kadavath2022language,he2021efficient}. Such a solution is also applicable to RA-LLMs, for example, FLARE~\cite{jiang2023active} dynamically triggers RAG if logits are lower than a specific threshold. 
\citet{tan2024small} introduce a more flexible model SlimPLM, which detects missing knowledge in LLMs with a slim proxy model, which functions to generate a ``heuristic answer''. The ``heuristic answer'' is used to assess the necessity of retrieval and facilitate the retrieval process for query rewriting when necessary. 

In traditional RAGs that rarely consider retrieval necessity, retrieval frequency (also called retrieval stride) is an important design aspect to determine the degree of using the retrieval in the generation, thereby greatly affecting the overall performance of RAG models~\cite{ram2023context}. Retrieval frequency controls how much to rely on the retrieval results, thereby affecting both the efficiency and effectiveness of the model. When the necessity of retrieval is not considered, retrieval frequency is often pre-defined and fixed, which have three common settings: one-time, every-n-token, and every-token. 
\textbf{One-time} retrieval invokes the retrieval function only once and tries to find all desired information in that one-time operation. One-time retrieval is usually operated at the beginning of the generation process, and then provides all retrieved documents to the generation models along with the original input, as applied in REALM~\cite{guu2020retrieval}. 
One-time retrieval is more suitable for the cases that the information needs in external databases are obvious to LLMs~\cite{jiang2023active}. 
However, for language tasks requiring long-form output such as open-domain summarization, the dependency among the tokens in the output is more important to be considered during the generation. 
In these cases, pre-retrieved documents (through one-time retrieval) might not be enough to support the generation of the whole sequence of output, which calls for in-generation retrieval operations. 
To this end, In-Context RALM~\cite{ram2023context} and RETRO~\cite{borgeaud2022improving} apply \textbf{every-n-token} retrieval in the process of generation for better augmentation. 
In comparison, kNN-LM~\cite{khandelwal2019generalization} adopts a more frequent retrieval strategy, which retrieves information for the prediction of \textbf{every token} during the generation. 
Overall, applying different frequencies of retrieval can impact both the effectiveness and efficiency of the whole RAG method. 
For example, more frequent retrieval leads to better performance but also increases the computing cost~\cite{ram2023context}. Choosing retrieval frequency is almost a trade-off between computing cost and performance. 

%% file: RAG/Training.tex
\section{RA-LLMs Training}
\label{sec:training}
Based on whether training is required or not, existing RAG methods can be categorized into two main classes: \textbf{train-free} approaches and \textbf{training-based} approaches. 
Training-free methods usually directly leverage the retrieved knowledge during inference time without introducing extra training by inserting the retrieved text into the prompt, which is computationally efficient. 
However, one potential challenge is that the retriever and generator components are not specifically optimized for downstream tasks, which could easily lead to sub-optimal utilization of the retrieved knowledge. 
To fully exploit the external knowledge, extensive methods are proposed to fine-tune the retriever and generator, thereby guiding large language models to effectively adapt and integrate retrieved information~\cite{sarto2022retrieval,wang2023shall,schick2024toolformer,zhu2024realm,shao2023enhancing,shi2023replug}.  

\begin{figure*}[t]
\centering
\includegraphics[width=1\linewidth]{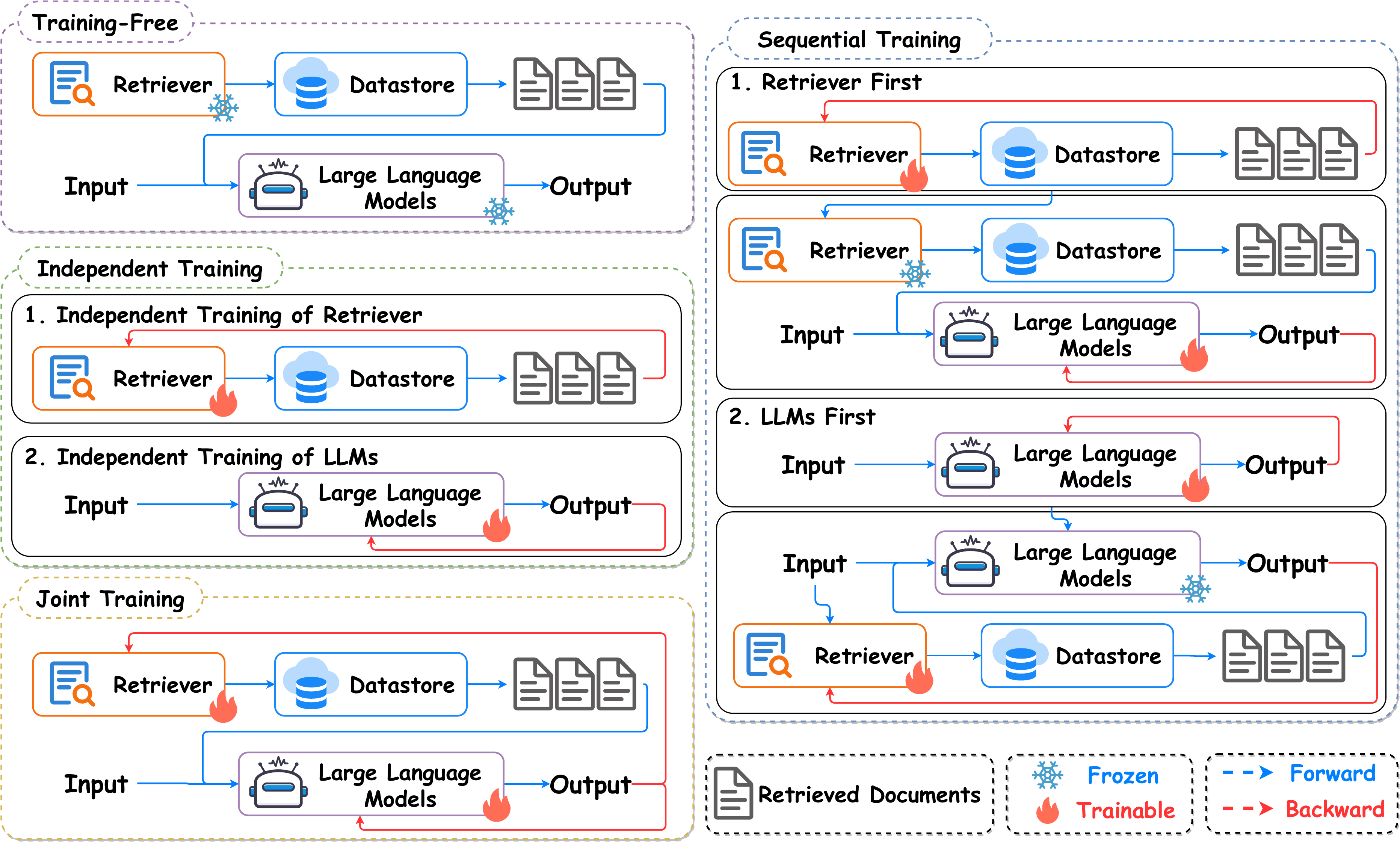}
\caption{An illustration of different training methods in Retrieval-Augmented Large Language Models  (RA-LLMs). 
Existing RA-LLMs approaches can be categorized into two classes: training-free approaches usually directly leverage retrieved information during the inference time by integrating the retrieved knowledge into the prompt, and training-based approaches fine-tune the retrieval and generator to enhance the generation performance. 
Based on the training strategies, training-based methods can be further categorized into three groups: independent training, where the retrieval and generator components are trained independently; sequential training, where they are trained sequentially; and joint training, where they are trained jointly.
}
\label{fig:training}
\end{figure*}

According to the training strategies, we categorize these training-based approaches into three classes: 
1) \textbf{Independent Training} approaches independently train each component in the RAG procedure, 
2) \textbf{Sequential Training} methods train one module first and freeze the well-trained component to guide the tuning process of the other part, 
and 3) \textbf{Joint Training} approaches train retriever and generator simultaneously. 
In the following section, we will comprehensively review the training-free, independent training, sequential training, and joint training methods. 
The comparison of these different training methods is depicted in Figure~\ref{fig:training}. 

\subsection{Training-free}\label{sec:inference_only}
With the huge number of parameters, LLMs have exhibited human-level intelligence and achieved promising prediction performance on various downstream tasks.
However, it is extremely challenging to frequently perform fine-tuning and update the knowledge stored in the model parameters~\cite{lewis2020retrieval} due to the considerable time and computational resources required. 
Recently, numerous studies have suggested enhancing LLMs with retrieval mechanisms, enabling them to dynamically acquire new knowledge from external sources without extra training processes (i.e., \emph{training-free})~\cite{izacard2021leveraging,jiang2023active,khattab2022demonstrate}, instead of relying solely on the implicit knowledge encoded in the model’s parameters. 
These approaches have shown significant performance improvement for various knowledge-intensive tasks, such as open-domain question answering~\cite{lewis2020retrieval}.
According to the different ways in which LLMs utilize retrieved information, we categorize these training-free methods into two categories: 1) \textbf{Prompt Engineering-based Methods} integrate retrieved knowledge into the original prompt directly, 
and 2) \textbf{Retrieval-Guided Token Generation Methods} retrieve information to calibrate the token generation process.

\subsubsection{Prompt Engineering-based Methods} 
As the LLMs' generation performance highly depends on the input query, numerous training-free RAG approaches employ external knowledge by refining the original prompts~\cite{jiang2023active,khattab2022demonstrate,li2023classification}. 
Specifically, the retrieved texts are usually used as contextual information and combined with the original prompt to guide the generation of LLMs~\cite{izacard2021leveraging,jiang2023active,khattab2022demonstrate,purwar2023keyword,li2023classification,wang2023knowledgpt,kim2023tree}. 
For example, 
In-Context RALM~\cite{ram2023context} keeps the LLM parameters unchanged and directly incorporates the retrieved document before the original prompt to augment the generation process. 
IRCoT~\cite{trivedi2023interleaving} interleaves chain-of-thought (CoT) generation and knowledge retrieval steps, enabling the retrieval of more relevant information for subsequent reasoning steps compared to standard retrieval methods that rely solely on the question as the query. 
Instead of retrieving knowledge from a large corpus, GENREAD~\cite{yu2023generate} first prompts a LLM to generate contextual documents based on the query, and then generate answers based on the given context and question. 
SKR~\cite{wang2023self} proposes guiding LLMs to determine whether they can answer a given question based on their internal knowledge, enabling flexible utilization of both internal and external knowledge by selectively calling the retriever. 
TOC~\cite{kim2023tree} first retrieves relevant knowledge for ambiguous questions and recursively constructs a tree structure by clarifying ambiguous questions into multiple disambiguate questions, which is further aggregated to generate long-form answers. 

\subsubsection{Retrieval-Guided Token Generation Methods} 
In addition to directly integrating external knowledge into the original prompt, the auxiliary information can be employed to adjust the token generation process. 
For example, KNN-KMs~\cite{khandelwal2019generalization} first retrieves $k$ most relevant contexts from the datastore based on the given query, and computes a neighbor distribution based on the distance. 
The output distribution is calibrated by interpolating the neighbor distribution and the original model's output distribution.  
Rest~\cite{he2023rest} is proposed to replace the parametric draft model with a non-parametric retrieval datastore and retrieves relevant tokens based on the current context for speculative decoding~\cite{chen2023accelerating,leviathan2023fast,sun2024spectr}. 



\subsection{Independent Training}
\label{sec:independent_training}
Independent training refers to training the retriever and LLMs as two entirely independent processes, in which there is no interaction between the retriever and the LLMs during the training process~\cite{karpukhin2020dense,zhou2022docprompting,lan2022copy}. 
Compared with training-free methods, the performance of the RAG-empowered models can be effectively enhanced by training LLMs to leverage the retrieved knowledge or retrievers to bridge the gap between information retrieval and language generation.
For the training of LLMs, the negative loglikelihood loss is the most representative training objective~\cite{radford2019language,touvron2023llama}, which aims to guide the LLMs to generate desired output based on the given input. 
Regarding the retriever, it can be categorized into two types: 1) Sparse retriever~\cite{ramos2003using,robertson2009probabilistic}, and 2) Dense retriever~\cite{lan2022copy,karpukhin2020dense,zhou2022docprompting}. 
The sparse retrievers usually exploit sparse features, e.g., word frequencies, to represent the documents and calculate the relevance scores based on task-specific metrics ~\cite{li2023empowering,ramos2003using, robertson2009probabilistic} such as TF-IDF and BM25.
As for the dense retrievers, deep neural networks are employed to encode the query and documents into dense representations, and then the inner product is usually used to calculate relevance scores and retrieve the relevant external knowledge. 
For example, DPR~\cite{karpukhin2020dense} adopts two independent BERT~\cite{devlin2018bert} networks to encode the query and passages respectively, and trains these models by utilizing contrastive learning. 
CoG~\cite{lan2022copy} proposes to train a prefix encoder and a phrase encoder for retrieval and reformulate the text generation as multiple copy-and-paste operations from existing source text collection.

\subsection{Sequential Training}
\label{sec:sequential_training}
Independent training is an efficient approach to exploit the external knowledge during the generation process since the retriever and generator can be trained offline and any off-the-shelf models can be utilized, avoiding extra training costs. 
To better enhance the synergy between the retriever and generator, several methods have been proposed to train the retriever and LLMs sequentially. 
In these sequential training methods, the process typically begins with the independent pre-training of either the retriever or the generator, after which the pre-trained module is fixed while the other module undergoes training. 
Note that various existing models (e.g., BERT~\cite{devlin2018bert,reimers2019sentence,khattab2020colbert}, CLIP~\cite{radford2021learning}, T5~\cite{raffel2020exploring}) can be directly employed as the fixed retriever and generator, thereby bypassing the first pertaining process. 
Compared to independent training, sequential training involves coordinated training of the retriever and generator, where the trainable module benefits from the assistance of the fixed module. 
Based on the training order between the retriever and generator, sequential training can be categorized into two classes: 1) \textbf{Retriever First}~\cite{sarto2022retrieval,wang2023shall,schick2024toolformer,zhu2024realm,asai2023self}, and 2) \textbf{LLMs First}~\cite{shi2023replug,wang2023learning,shao2023enhancing}. 

\subsubsection{Retriever First} These methods first train the retrieval model and then fix it. LLMs are then trained by utilizing the retrieved knowledge. 
For instance, RETRO~\cite{borgeaud2022improving} adopts the BERT model that is pre-trained independently as the retriever, and an encoder-decoder architecture is trained to integrate retrieval chunks into the model's predictions. 
RALMs~\cite{yoran2023making} adopts Google Search and the open-source COLBERTV2~\cite{khattab2020colbert} as the pre-trained retriever and fine-tunes the LLM to effectively leverage the retrieved passages. 
ITER-RTGEN~\cite{ren2023retrieve} utilizes the pre-trained S-BERT~\cite{reimers2019sentence} as the retriever and introduces an adaptive hybrid retrieval strategy for retrieving demonstrations. Additionally, it leverages T5~\cite{raffel2020exploring} as the generator, which undergoes further fine-tuning based on the target label and input combining the original prompt with retrieved demonstrations.
SMALLCAP~\cite{ramos2023smallcap} proposes using the CLIP~\cite{radford2021learning}, which is a powerful pre-trained multi-modal network, to encode the input image and the textual data of the external datastore and retrieve the most relevant items based on the cosine similarity. A cross-attention layer is trained and GPT-2~\cite{radford2019language} is used as the decoder to produce captions.


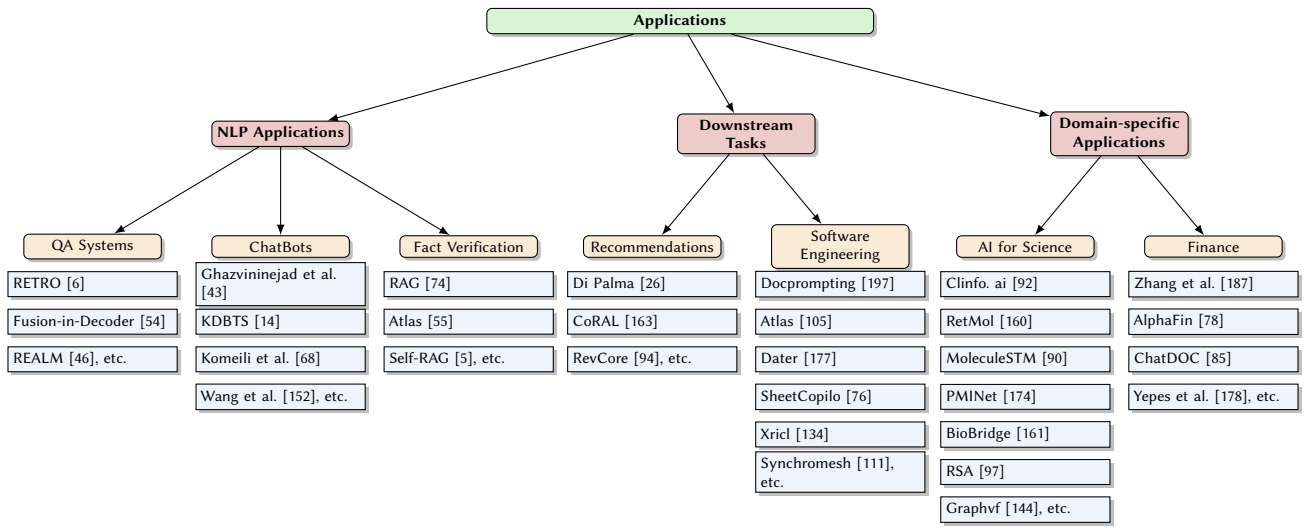
\begin{figure*}[t]
\centering
\scriptsize
\begin{tikzpicture}
[node distance=0.5cm,
  level 1/.style={sibling distance=25mm},
  edge from parent/.style={->,draw},
  >=latex][t]
\node[root] {\textbf{Applications}}
  child {node [level 2, xshift=-80pt] (b1) {\textbf{NLP Applications}}
    child {node [level 3] (c1) {QA Systems}}
    child {node[level 3] (c2) {ChatBots}}
    child {node[level 3] (c3) {Fact Verification}}
  }
   child {node [level 2, xshift=25pt] (b2) {\textbf{Downstream Tasks}}
    child {node [level 3] (c4) {Recommendations}}
    child {node[level 3] (c6) {Software \\Engineering}}
  }
  child {node [level 2, xshift=95pt] (b3) {\textbf{Domain-specific Applications}}
    child {node[level 3] (c5) {AI for Science}}
 child {node[level 3] (c7) {Finance}}
  };

\begin{scope}[every node/.style={level 4}]
\node [below of = c1] (c10) {RETRO~\cite{borgeaud2022improving}};
\node [below of = c10 ] (c11)
{Fusion-in-Decoder~\cite{izacard2021leveraging}};
\node [below of = c11 ] (p11)
{REALM~\cite{guu2020retrieval}, etc.};

\node [below of = c2] (c20) {\citet{ghazvininejad2018knowledge}};
\node [below of = c20 ] (c21)
{KDBTS~\cite{chen2020bridging}};
\node [below of = c21] (c22) {\citet{komeili2021internet}};
\node [below of = c22 ] (p28)
{\citet{wang2023search}, etc.};

\node [below of = c3 ] (c31) {RAG~\cite{lewis2020retrieval}};
\node [below of = c31 ] (c31)
{Atlas~\cite{izacard2023atlas}};
\node [below of = c31 ] (p311) {Self-RAG~\cite{asai2023self}, etc.};

On the other hand, \citet{yepes2024financial} propose a document chunking method based on structure instead of chunking based on paragraphs, further improving the quality of RA-LLMs outputs.
\node [below of = c4 ] (c41) {\citet{di2023retrieval}};
\node [below of = c41 ] (c42)
{CoRAL~\cite{wu2024coral}};
\node [below of = c42 ] (c43) {RevCore~\cite{lu2021revcore}, etc.};

\node [below of = c5 ] (c51) {Clinfo. ai~\cite{lozano2023clinfo}};
\node [below of = c51 ] (c53) {RetMol~\cite{wang2022retrieval}};
\node [below of = c53 ] (c54) {MoleculeSTM~\cite{liu2023multi}};
\node [below of = c54 ] (c55)
{PMINet~\cite{yang2023prompt}};
\node [below of = c55 ] (c56) {BioBridge~\cite{wang2023biobridge}};
\node [below of = c56 ] (c57) 
{RSA~\cite{ma2023retrieved}};
\node [below of = c57 ] (c58) {Graphvf~\cite{sun2023graphvf}, etc.};

\node [below of = c6 ] (c61) {Docprompting~\cite{zhou2022docprompting}};
\node [below of = c61 ] (c62)
{Atlas~\cite{nashid2023retrieval}};
\node [below of = c62 ] (c63) {Dater~\cite{ye2023large}};
\node [below of = c63 ] (c64) {SheetCopilo~\cite{li2024sheetcopilot}};
\node [below of = c64 ] (c65)
{Xricl~\cite{shi2022xricl}};
\node [below of = c65 ] (c66) {Synchromesh~\cite{poesia2022synchromesh}, etc.};

\node [below of = c7 ] (c71) {\citet{zhang2023enhancing}};
\node [below of = c71 ] (c72)
{AlphaFin~\cite{li2024alphafin}};
\node [below of = c72 ] (c73) {ChatDOC~\cite{lin2024revolutionizing}};
\node [below of = c73 ] (c74) {\citet{yepes2024financial}, etc.};

\end{scope}

\end{tikzpicture}
\caption{A summary of applications of RA-LLMs categorized by \emph{NLP applications}, \emph{downstream tasks}, and \emph{domain-specific application}. Specifically, NLP applications include QA systems, ChatBots, and fact verification; downstream tasks include recommendations and software engineering; and domain-specific applications include  AI for Science and Finance.
} 
\label{fig:applications}
\end{figure*}

\subsubsection{LLMs First} Similarly, it can also pre-train LLMs first, and then tune the retriever under the supervision of the well-trained LLMs. 
For example, DKRR~\cite{izacard2021distilling} shows that attention scores from a sequence-to-sequence model can indicate the document's relevance. Therefore, they propose to leverage the attention scores of a reader model to produce synthetic labels to train the retriever. 
AAR~\cite{yu2023augmentation} proposes using a small language model to generate the supervised signal for training retrievers. The well-trained retriever can be further leveraged to enhance the performance of black-box LLMs.
RA-DIT~\cite{lin2023ra} first fine-tunes the LLMs to enhance their ability to leverage retrieved knowledge, and then train the retriever to better align its output with LLMs. 
UPRISE~\cite{cheng2023uprise} proposes a lightweight method to enhance the zero-shot performance of LLMs in unseen tasks by introducing a prompt retriever. A frozen LLM is employed to guide the fine-tuning process of the prompt retriever, and this retriever then retrieves prompts for different tasks with various LLMs during inference. 


\subsection{Joint Training}
\label{sec:joint_training}

Joint training methods~\cite{zhong2022training,kang2023knowledge,li2023structure,xu2023retrieval,hu2023reveal,cheng2024lift} employ the end-to-end paradigm to optimize the retriever and generator simultaneously. Instead of training each module sequentially, joint training methods effectively enhance the retriever's ability to locate external knowledge for generation and the generator's capacity to effectively leverage the retrieved information. 
For instance, RAG~\cite{lewis2020retrieval} minimizes the negative loglikelihood to jointly train the retriever and generator. 
REALM~\cite{guu2020retrieval} adopts a similar training paradigm to that of RAG~\cite{lewis2020retrieval}, and Maximum Inner Product Search (MIPS)~\cite{ram2012maximum,chen2019deep,shen2015learning,ding2020bilinear} technique is used to locate the most relevant documents. 
To employ MIPS, all external documents are embedded first and a search index is produced for each embedding. 
An asynchronous index updating strategy~\cite{guu2020retrieval,izacard2023atlas,siriwardhana2023improving,huang2023raven} is proposed to refresh the index every several hundred training steps to avoid time consumption of re-indexing all documents.



%% file: RAG/Applications.tex
\section{Applications}\label{sec:applications}
In this section, we will introduce some representative applications of retrieval-augmented large language models (RA-LLMs). 
To provide a clear overview of the applications of RA-LLMs, we will review them from three perspectives: \emph{NLP applications}, \emph{downstream tasks}, and \emph{domain-specific applications}. 
The studies mentioned in this section are summarized and categorized in Figure~\ref{fig:applications}.

\subsection{NLP Applications}
Due to the intrinsic capability in text generation, RA-LLMs have various applications in the NLP field, such as Question Answer (QA) Systems, ChatBot, and Fact Verification.

\subsubsection{QA Systems}
QA Systems aim to provide precise answers to user's queries.
However, even when trained on extensive data, these systems may lack the latest information or specific domain knowledge that is not included in their training data~\cite{izacard2021leveraging,liu2022uni}. 
To address this limitation, the integration of RA-LLMs has played a crucial role in advancing the capabilities of QA systems by enhancing their ability to retrieve and synthesize relevant information~\cite{borgeaud2022improving,izacard2021leveraging}.
Specifically, RA-LLMs can provide coherent and contextually relevant answers by leveraging their retrieval component to access a vast knowledge base.
For example, REALM~\cite{guu2020retrieval} integrates a knowledge retriever that can retrieve information from a large corpus during pre-training, fine-tuning, and inference. 
This approach allows REALM to effectively retrieve from a vast knowledge corpus, thereby improving the accuracy of its responses. 
Similarly, Fusion-in-Decoder~\cite{izacard2021leveraging} retrieves passages from support documents and then fuses them with questions to generate the answer, achieving higher accuracy.
In addition, ~\citet{borgeaud2022improving} indicate that the quality of the answers may rely more on the output of retrieval.

\subsubsection{ChatBot}
ChatBot is designed to interact with users in a natural and conversational manner~\cite{liu2020does}.
Different from the QA system, ChatBot focuses on maintaining a coherent and contextually rich conversation with the user.
To enhance these capabilities, recent methods focus on integrating RA-LLMs~\cite{komeili2021internet,zhang2019grounded,kang2023knowledge} for its ability to augment the ChatBot with relevant external knowledge, facilitating more engaging and context-rich interactions with users.
For example, some studies~\cite{ghazvininejad2018knowledge,chen2020bridging} retrieve relevant knowledge from static databases (e.g., a Wikipedia dump) to augment conversation. 
~\citet{komeili2021internet} propose retrieving information from the internet search to further augment conversation performance.
Considering the dynamic nature of knowledge in the world, another model~\cite{wang2023search} further accesses large amounts of dynamic information in search engines to generate responses.

\subsubsection{Fact Verification} 
Fact Verification is a critical task in verifying the accuracy and reliability of information. 
With the need for trusted evidence, RA-LLMs are being utilized to enhance the capabilities of fact verification~\cite{lewis2020retrieval,izacard2023atlas,lewis2020retrieval}.
~\citet{lewis2020retrieval} first propose retrieval of external knowledge to augment a range of knowledge-intensive tasks including fact verification.
On the other hand, Atlas~\cite{izacard2023atlas} examines the performance of the RA-LLMs for fact verification under few-shot learning.
Recently, Self-RAG~\cite{asai2023self} has greatly made a notable impression by incorporating a self-reflective mechanism. Specifically, Self-RAG reflects on whether retrieved information is helpful and judges the reliability of retrieved information, thereby greatly improving the verification accuracy.

\subsection{Downstream Tasks}
In addition to NLP applications, RA-LLMs can also be applied to various downstream tasks, such as recommendations and software engineering.

\subsubsection{Recommendations}
Recommender systems play an important role in modeling users' preferences and providing personalized recommendations~\cite{zhang2024linear,wang2024rethinking,fan2019graph,zhao2024recommender,fan2020graph,fan2022graph}.
Recently, RA-LLMs have demonstrated great potential in providing personalized and contextually relevant recommendations by integrating retrieval and generation processes~\cite{di2023retrieval,wu2024coral,lu2021revcore}.
For example, ~\citet{di2023retrieval} proposes a simple retrieval-augmented recommendation model, that leverages knowledge from movie or book datasets to enhance recommendations.
Additionally, ~\citet{lu2021revcore} further retrieval from the reviews to enrich item information in recommender systems.
CoRAL~\cite{wu2024coral} utilizes reinforcement learning to retrieve collaborative information from the dataset and align it with semantic information for more accurate recommendations.

\subsubsection{Software Engineering}
The rise of RA-LLMs has influenced many aspects of software engineering~\cite{zhou2022docprompting,nashid2023retrieval,ye2023large}.
For example, some studies propose the retrieval-augmented generation paradigm for code generation~\cite{zhou2022docprompting} and program repair~\cite{nashid2023retrieval}.
Similarly, ~\citet{parvez2021retrieval} retrieve top-ranked codes or summaries from the codebase and aggregate them with input to enhance code generation and summarization.
In addition, RA-LLMs show potential in tabular
data processing~\cite{ye2023large,li2024sheetcopilot} and Text-to-SQL semantic parsing~\cite{shi2022xricl,poesia2022synchromesh}.

\subsection{Domain-specific Applications}
RA-LLMs have been widely adopted for various domain-specific tasks, such as AI for Science and Finance.

\subsubsection{AI for Science}
RA-LLMs have proven to be beneficial for the realms of science, such as molecular and protein. \textbf{Molecules} include identifying the molecule's property and predicting new molecules, thereby favoring drug discovery.
Currently, some RA-LLMs have been applied to molecules by integrating retrieval of molecule structure and biomedical entities like protein, molecule, and disease ~\cite{wang2022retrieval,liu2023multi,yang2023prompt,wang2023biobridge}, etc.
\citet{wang2022retrieval,li2023empowering} propose retrieval-based frameworks by retrieving from the database to guide molecule generation.
\citet{liu2023multi} introduce a multi-modal molecule structure-text model by retrieving textual knowledge from a large-scale dataset for molecular property prediction.
In addition, RA-LLMs also significantly influence \textbf{Protein} representation and generation~\cite{sun2023graphvf,ma2023retrieved}. 
For instance, RSA~\cite{ma2023retrieved} queries protein sequences associated with a collection of structurally or functionally similar sequences in the database to enhance protein representations.
Furthermore, ~\citet{lozano2023clinfo} present a clinical QA system based on retrieving published review articles.

\subsubsection{Finance}
In the highly data-driven and information-intensive field of finance, RA-LLMs have proved to be a significant technology for enhancing decision-making~\cite{zhang2023enhancing,yepes2024financial,li2024alphafin}.
For example, \citet{zhang2023enhancing} retrieve financial information from external sources, such as news platforms (e.g., Bloomberg and Reuters) and social media platforms (e.g., Twitter, Reddit), to combine with the original query to enhance the precision of financial sentiment analysis.
In addition, financial QA is another primary task of financial analysis, which usually extracts relevant knowledge from financial documents. As professional documents are usually stored in PDF format, \citet{lin2024revolutionizing} introduces a PDF parser combined with RA-LLMs to retrieve knowledge from financial reports.
On the other hand, \citet{yepes2024financial} propose a document chunking method based on structure instead of chunking based on paragraphs, further improving the quality of RA-LLMs outputs.

%% file: RAG/Future_work.tex
\section{Future Challenges and Opportunities}
\label{sec:future}
Since the studies of RA-LLMs are still in the early stage, we present some potential research directions that can be explored in the future in the field of RA-LLMs.

\noindent \textbf{Trustworthy RA-LLMs}. 
The essential objective of developing RAG-empowered LLMs is to enhance the capability of the language models, thereby benefiting users and society by alleviating redundant and meaningless labor, increasing conveniences, and spurring social progress.
However, recent research indicates that RA-LLMs can be maliciously and unintentionally manipulated to make unreliable decisions and harm humans~\cite{deng2024pandora,zou2024poisonedrag}, which may have serious consequences in safety-critical scenarios~\cite{liu2021trustworthy,fan2022comprehensive,fan2021attacking,chen2023fairly,chen2022knowledge}. In addition, private retrieval database has a risk of leakage, raising concerns regarding the privacy of RA-LLMs~\cite{zeng2024good}. 
Therefore, developing trustworthy RA-LLMs is of paramount importance as it can significantly mitigate the potential negative impacts of LLMs technology and provide people with powerful AI models that can be fully trusted. 
To be specific, the ideal trustworthiness in RA-LLMs systems should possess the following characteristics: 1) \textbf{robustness}, 2) \textbf{fairness}, 3) \textbf{explainability}, and 4) \textbf{privacy}. 
For example, \textbf{robustness} means a trustworthy RA-LLMs system should be robust against malicious or inadvertent perturbations introduced by attackers. 
\textbf{Fairness} indicates a trustworthy RA-LLMs system ought to avoid discrimination during the decision-making process.
\textbf{Explainability} requires a complete understanding of the intrinsic workings of RA-LLMs systems, i.e., the predictions of RA-LLMs systems are explainable and transparent. 
\textbf{Privacy} entails safeguarding the safety of this private information housed within the datastore when establishing trustworthy RA-LLMs systems.

\noindent \textbf{Multi-Lingual RA-LLMs}. 
The ability of leveraging knowledge from multiple languages can greatly enhance the capabilities of retrieval-augmented language models. 
As the world becomes increasingly interconnected, there is a growing need for AI systems that can understand and communicate across different languages. 
By incorporating multilingual knowledge retrieval and generation, these models can access and synthesize information from diverse linguistic sources, enabling more comprehensive and nuanced understanding and generation capabilities. 
Additionally, multilingual models can facilitate cross-cultural communication and knowledge sharing and breaking down language barriers, thereby bringing convenience to people across different regions of the world, especially those in areas with minority languages~\cite{kabra2023multi,li2023classification}. 
For example, users from countries with less prevalent languages can utilize abundant English and Chinese corpora for knowledge retrieval, enhancing the performance of large language models in downstream tasks.

\noindent \textbf{Multi-modal RA-LLMs}.
Multi-modal retrieval-augmented generation extends the knowledge sources beyond text to include various data modalities such as images, videos, and audio. 
By integrating various modalities, LLMs can leverage richer contextual information than single-modal RAG and develop a more comprehensive understanding of users' needs, bringing precise, fine-grained, and high-quality generation. 
For instance, an image or video can provide valuable visual cues that complement textual information, leading to more precise language generation~\cite{zhu2024realm,hu2023reveal}. 
By fusing multiple modalities, multi-modal RA-LLMs can develop a more comprehensive understanding of the world, leading to more accurate and insightful outputs, benefiting a wide range of domains, including healthcare~\cite{zhu2024realm}, drug discovery~\cite{shtar2021multimodal}, molecular analysis~\cite{liu2023multi,andrews2022multi}, etc. 

\noindent \textbf{Quality of External Knowledge}.
As a commonly used datastore in current RAG systems, Wikipedia~\cite{zhu2024realm,karpukhin2020dense} serves as a vast repository of external textual knowledge used to augment the generation process, which contains millions of articles covering various disciplines.
However, the reliability and accuracy of individual articles within Wikipedia vary significantly, and the introduction of some texts that deviate from facts might even mislead the model's generation process. 
Therefore, it is crucial to enhance the quality of the external knowledge corpus and mitigate the negative impact of low-quality knowledge on the performance of LLMs.
By enhancing the quality of the external knowledge and tailing robust mechanisms by filtering out low-quality or unreliable information, the RA-LLM systems might produce more accurate, reliable outputs, thereby improving their effectiveness in various real-world applications.

%% file: RAG/Conclusion.tex
\section{Conclusion}
\label{sec:conclusion}

Retrieval-augmented generation (RAG), a cutting-edge AI technique, has achieved remarkable success across various applications, including recommendation, molecule generation, protein representation, and software engineering, owing to the potent capabilities of retrieval in providing supplementary information to enhance generation performance. 
Recently, increasing efforts have been made to alleviate the limitations of large language models (LLMs), such as hallucination and out-of-date internal knowledge, by leveraging retrieval to provide the latest auxiliary information and teaching LLMs to harness the retrieved external knowledge. 
With the rapid advancements in retrieval-augmented large language models (RA-LLMs), there is a pressing need for a comprehensive and systematic overview. 
To bridge this gap, in this paper, we comprehensively review the RA-LLMs from the perspectives of morel architecture, training strategy, and application area, providing researchers with an in-depth understanding. 
Moreover, since the studies of RA-LLMs are still in the early stage, we also discuss the current limitations and several potential research directions for future research.